EURASIP Journal on Image and Video Processing
a SpringerOpen Journal

## RESEARCH                                                    Open Access

# Face recognition using color local binary pattern from mutually independent color channels

Gholamreza Anbarjafari

### Abstract

In this article, a high performance face recognition system based on local binary pattern (LBP) using the probability distribution functions (PDFs) of pixels in different mutually independent color channels which are robust to frontal homogenous illumination and planer rotation is proposed. The illumination of faces is enhanced by using the state-of-the-art technique which is using discrete wavelet transform and singular value decomposition. After equalization, face images are segmented by using local successive mean quantization transform followed by skin color-based face detection system. Kullback–Leibler distance between the concatenated PDFs of a given face obtained by LBP and the concatenated PDFs of each face in the database is used as a metric in the recognition process. Various decision fusion techniques have been used in order to improve the recognition rate. The proposed system has been tested on the FERET, HP, and Bosphorus face databases. The proposed system is compared with conventional and the state-of-the-art techniques. The recognition rates obtained using FVF approach for FERET database is 99.78% compared with 79.60 and 68.80% for conventional gray-scale LBP and principle component analysis-based face recognition techniques, respectively.

## Introduction

Face recognition has been one of the most interesting research topics for over the past half century. During this period, many methods such as principle component analysis (PCA), linear discriminant analysis (LDA), independent component analysis (ICA), etc., have been introduced [1-5]. Many of these methods are based on gray-scale images; however, color images are increasingly being used since they add additional biometric information for face recognition [6-8]. As reported by Demirel and Anbarjafari [6,8], color probability distribution functions (PDFs) of a face image can be considered as the signature of the face, which can be used to represent the face image in a low-dimensional space. It is known that PDF of an image is a normalized version of an image histogram [9]. PDF recently has been used in many applications of image processing such as object detection, face localization, and face recognition [6,8-12].

One of the most important steps in a face recognition system is face segmentation. There are various methods for segmentation of the faces such as skin color-based face segmentation [13,14], Viola–Jones [15] face detection system, local successive mean quantization transform (SMQT)-based face detection [16,17]. In this study, we are using local SMQT-based face segmentation followed by skin color-based face segmentation. This procedure will reduce the effect of background on the rectangle-shape segmented face image.

In this article, the PDF-based face recognition will be studied analytically and then LBP will be used in order to boost the recognition performance. Also in this article, instead of experimentally choosing PDFs of *HSI* and *YCbCr* color channels [6,8], analytically specific color channels have been selected. Furthermore, analytical studies of false acceptance rate (FAR) and false rejection rate (FRR) analysis are included in the third section. The head pose (HP) face database [18] with 15 subjects, a subset of 50 subjects from the FERET [19] database with faces containing varying poses changing from −90° to +90° of rotation around the vertical axis passing through the neck (the same subset

Correspondence: sjafari@ciu.edu.tr
Department of Electrical and Electronic Engineering, Cyprus International University, Lefkoşa, KKTC, Mersin 10, Turkey

Springer

as Demirel and Anbarjafari used in [6,8]), and Bosphorus face database [20] with 105 subjects with varying frontal illuminations, poses, expressions, and occlusions were used to test the proposed system.

## Facial images pre-processing

### Image illumination enhancement

In many image processing applications, the general histogram equalization (GHE) is one of the simplest and most effective primitives for contrast enhancement [21], which attempts to produce an output histogram that is uniform [22]. One of the disadvantages of the GHE is that the information laid on the histogram or PDF of the image will be lost. Demirel and Anbarjafari [6] have showed that the PDF of face images can be used for face recognition, hence preserving the shape of PDF of the image is of vital importance. Therefore, GHE is not a suitable technique for illumination enhancement of face images. Also, it is known that GHE often produces unrealistic effects in images. After the introduction of GHE, researchers came out with better technique which deals with equalization of portion of the image at a time, called local histogram equalization (LHE). LHE can be expressed as follows: GHE can be applied independently to small regions of the image. Most small regions will be very self-similar. If the image is made up of discrete regions, most small regions will lie entirely within one or the other region. If the image has more gradual large-scale variation, most small regions will contain only a small portion of the large-scale variation.

However, the contrast issue is yet to be improved and even these days many researchers are proposing new techniques for image equalization. DHE is obtained from dynamic histogram specification [23] which generates the specified histogram dynamically from the input image.

Demirel and Anbarjafari [6,24] developed singular value decomposition (SVD)-based image equalization (SVE) technique, which is based on equalizing the singular value (SV) matrix obtained by SVD. Since an image can be considered as a numeric matrix, SVD of an image, $A$, can be written as follows

$$A = U_A \Sigma_A V_A^T \tag{1}$$

where $U_A$ and $V_A$ are orthogonal square matrices known as hanger and aligner, respectively, and $\Sigma_A$ matrix contains the sorted SVs on its main diagonal. The idea of using SVD for image equalization comes from this fact that $\Sigma_A$ contains the intensity information of the given image [25]. The objective of SVE [24] is to equalize a low-contrast image in such a way that the mean moves towards the neighborhood of 8-bit mean gray value 128 in the way that the general pattern of the PDF of the image is preserved. Demirel and Anbarjafari [6] used SVD to deal with the illumination problem in their proposed face recognition system. SVE can be described in the following way: the ratio of the largest SV of the generated normalized matrix over a normalized image. This coefficient can be used to regenerate an equalized image. This task is eliminating the illumination problem. It is important to mention that techniques such as DHE or SVE are preserving the general pattern of the PDF of an image.

The proposed method is robust to the frontal homogenous illumination changes and this robustness is achieved by using a state-of-the-art technique which is based on discrete wavelet transform (DWT) and SVD [26]. The two-dimensional wavelet decomposition of an image is performed by applying the one-dimensional DWT along the rows of the image first, and then the results are decomposed along the columns. This operation results in four decomposed sub-band images refer to low-low (LL), low-high (LH), high-low (HL), and high-high (HH). The frequency components of those sub-band images cover the frequency components of the original image. DWT is used to separate the input image into different frequency sub-bands, where LL sub-band concentrates the illumination information. The method is benefiting from the fact that an intensity image (an 8-bit intensity image in a given color channel), $A$, can be decomposed into multiplication of three matrices, by using SVD. The first SV has the highest impact on the illumination of the image, hence updating this SV will directly effect on the illumination of $A$. Also it is known that the equalized image by using GHE will result into a visually good looking image [27]. Furthermore, in wavelet domain the illumination information is laid in the LL sub-band. By knowing this information, the method is modifying the SVs by using a correction coefficient, $\zeta$:

$$\zeta = \frac{\max(\Sigma_{LL_{\hat{A}}})}{\max(\Sigma_{LL_A})} \qquad (2)$$

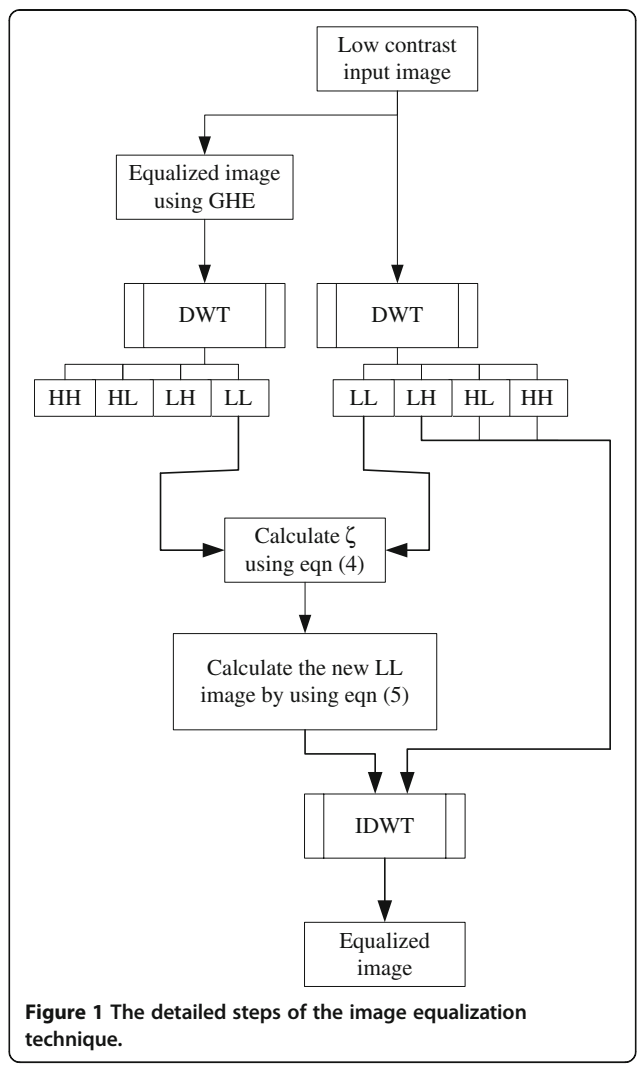

**Figure 1 The detailed steps of the image equalization technique.**

Then, the equalized image is being reconstructed by using the following equation:

$$\begin{aligned} \overline{\Sigma}_{LL_A} &= \zeta\, \Sigma_{LL_A} \\ \overline{LL_A} &= U_{LL_A}\overline{\Sigma}_{LL_A}V_{LL_A} \\ \overline{A} &= IDWT\left(\overline{LL_A}, LH_A, HL_A, HH_A\right) \end{aligned} \qquad (3)$$

SVD is a computationally complex operation. As it is shown in Equation (2) only the highest (the first) SV is used. Also from elementary algebra, it is known that for a matrix, the highest SV is obtained by calculating its norm. Thus, $\zeta$ can be calculated by

$$\zeta = \frac{\|LL_{\hat{A}}\|}{\|LL_A\|} \qquad (4)$$

Then the equalized image is being reconstructed by using the following equation

$$\begin{aligned} \overline{LL_A} &= \zeta LL_A \\ \overline{A} &= IDWT\left(\overline{LL_A}, LH_A, HL_A, HH_A\right) \end{aligned} \qquad (5)$$

Figure 1 shows the block diagram of the proposed technique. One can implement the algorithm in different color channel in order to enhance the illumination of the color images.

Figure 2 shows the visual effect of the DWT-based illumination enhancement on a face image with three different illuminations from OULU face database [28].

## Face localization and segmentation

A face is naturally recognizable by a human regardless of its many point of variation such as skin tone, facial hair, etc. Face detection is a required first step in face recognition systems [16,29]. The most straight forward variety of face localization is the detection of a single face at a known scale and orientation, which is yet a non-trivial problem. Efficient fast face detection is an impressive goal, which is subject to face tracking that required no knowledge of previous frames [30]. Another reason that face detection is an important research issue is its role as a challenging case of a more general problem, object detection.

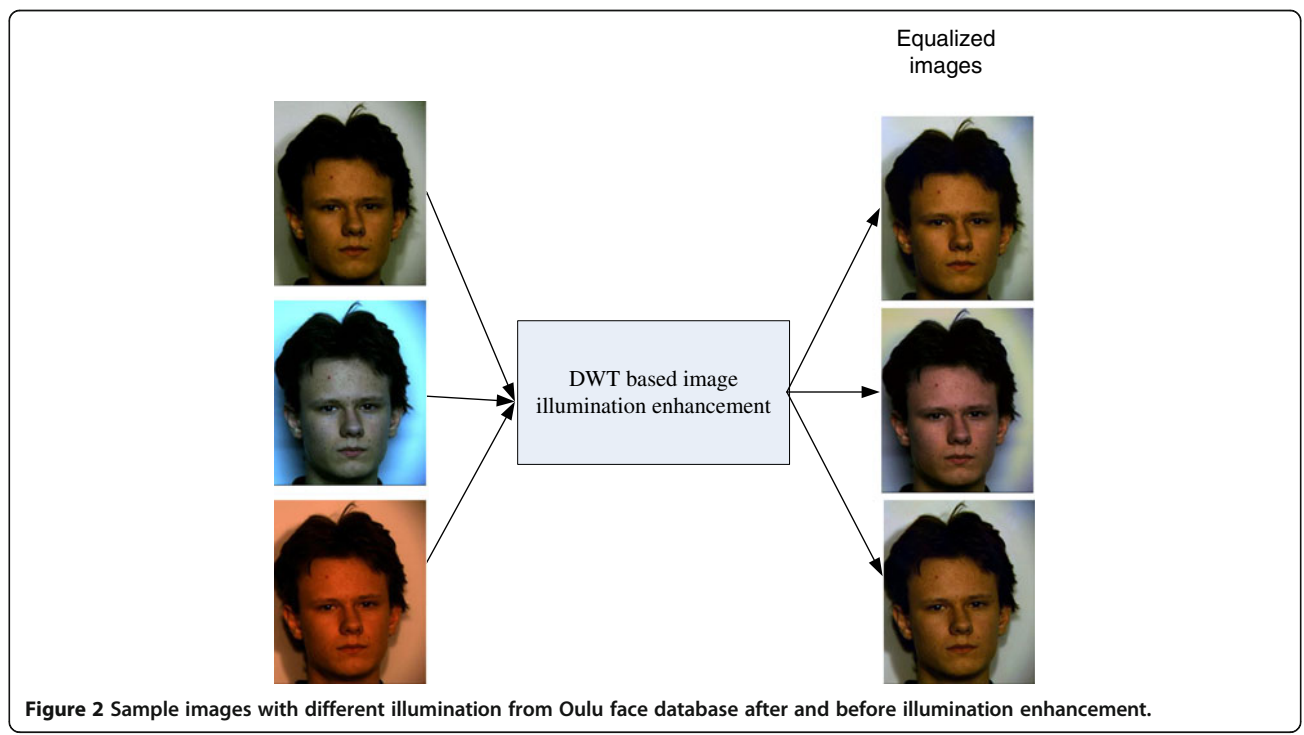

**Figure 2 Sample images with different illumination from Oulu face database after and before illumination enhancement.**

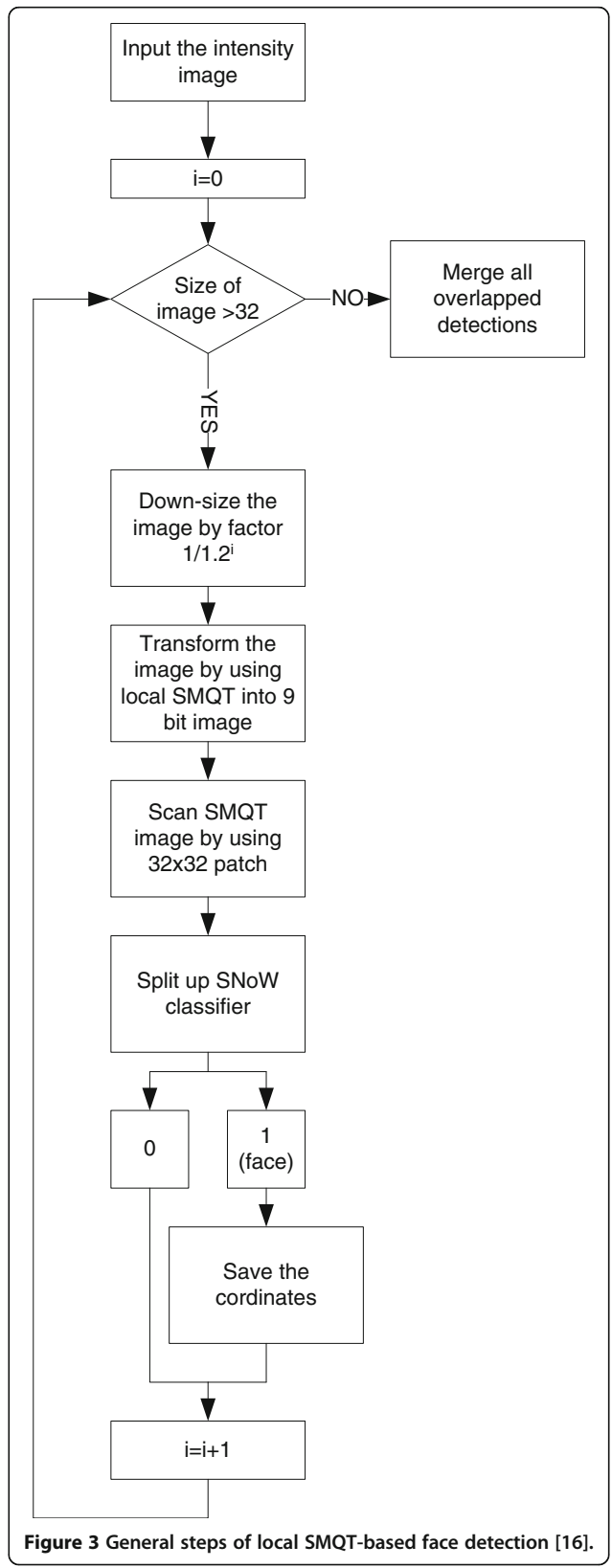

**Figure 3** General steps of local SMQT-based face detection [16].

Skin is a widely used feature in human image processing with applications ranging from face detection [13] and person tracking [31] to content filtering [32]. Human skin can be detected by identifying the presence of skin color pixels. Many methods have been proposed for achieving this. Chai and Ngan [14] modeled the skin color in *YCbCr* color space. In their technique, pixels are classified into "skin" and "non-skin" by using four threshold values, which form a rectangular region in *CbCr* space.

All color spaces such as *RGB*, *HIS*, and *YCbCr* can be used for face recognition [33]. The advantage of using *HSI* color space is its independence of knowledge of the exact percentage of red, green, or blue. Many applications such as machine vision use *HSI* color space in identifying the color of different objects.

Kjeldson and Kender [34] stated a color preference in *HSI* color space to distinguish skin regions from other segments. Skin color classification in *HSI* color space is based on hue and saturation values.

In [35], the threshold for hue and saturation has been modified from previous work [34] by using 2,500 face samples taken from the FERET dataset, the HP face database, and the Essex University face database. According to these samples, the thresholds for hue and saturation color channels are updated to satisfy the following threshold.

$$(H < 0.17 \quad OR \quad H > 0.63 \quad AND \quad S > 0.1 \qquad (6)$$

In order to eliminate the illumination effect from the input images, the intensity component of an image in *HSI* color space has been equalized.

A different approach to separating faces and non-faces in image space is proposed by Osuna et al. [36] and later developed and modified by Romdhani et al. [37]. Both are based on support vector machines (SVM) [38]. The key to the SVM model is the choice of a manifold that separates the face set from the non-face set. Romdhani et al. had chosen a hyperplane which maximizes minimum distance on either side. Romdhani et al. worked further on reducing the vector set in order to improve performance.

In this article, the proposed face recognition system uses local SMQT-based face detection followed by skin tone-based face localization. The SMQT can be considered as an adjustable trade-off between the number of quantization levels in the result and the computational load [39]. Local is defined to be the division of an image into blocks with a predefined size. Let *x* be a pixel of local *D*, and let's have the SMQT transform as follows

$$SMQT_L : D(x) \rightarrow M(x) \qquad (7)$$

where *M*(*x*) is a new set of values which are insensitive to gain and bias [39]. These two properties are desired for the formation of the intensity image which is a product of reflection and illumination. A common approach to separate the reflection and illumination is based on this assumption that illumination is spatially smooth so that it can be taken as a constant in a local area. Therefore, each local pattern with similar structure will yield the similar SMQT features for a specified level, $L$. The spare network of winnows (SNoW) learning architecture is also employed in order to create a look-up table for classification [40].

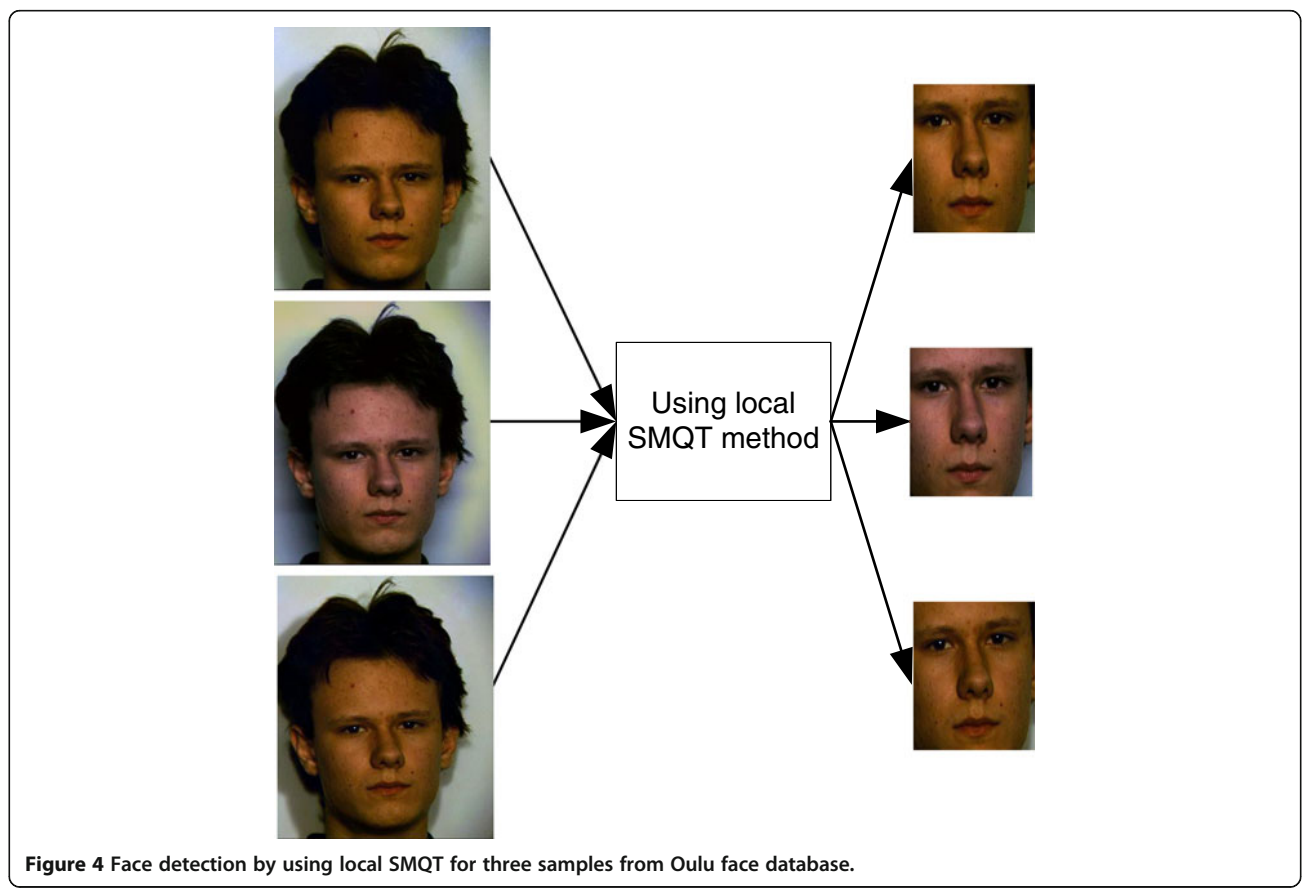

**Figure 4 Face detection by using local SMQT for three samples from Oulu face database.**

As Nilsson et al. [39] proposed, in order to scan an image for faces, a patch of $32 \times 32$ pixels is used and also the image is downscaled and resized with a scale factor to enable the detection of faces with different sizes. The choice of the local area and the level of the SMQT are vital for successful practical operation. The level of the transform is also important in order to control the information gained from each feature. As reported in [39], the $3 \times 3$ local area and level $L = 1$ are used to be a proper balance for the classifier. The face and non-face tables are trained in order to create the split up SNoW classifier. Overlapped detections are disregarded using geometrical locations and classification score. Hence, given two detections overlapping each other, the detection with the highest classification score is kept and the other one is removed. This operation is repeated until no overlapping detection is found.

The block diagram of the local SMQT-based face detection is shown in Figure 3.

Figure 4 shows the images of the three faces from Oulu face database enhanced by using DWT + SVE illumination enhancement and the segmented faces by using local SMQT.

The local SMQT-based face detector software has several advantages which promote us in order to use it in this study, which are

– it is fast and very accurate;
– it is a state-of-the-art technique;
– if the input image does not have a face image, there will be no output, therefore in the proposed face recognition system there will be no issue of having a noise image, whose PDF is the same as a face image, as an input.

The output of the local SMQT-based face recognition is that the face image is being cropped in a rectangle frame; hence, some part of background or hair is also included in the frame of the segmented face image. In order to reduce this effect, the output of the local SMQT face detector [17] is passed through a skin color-based face segmentation

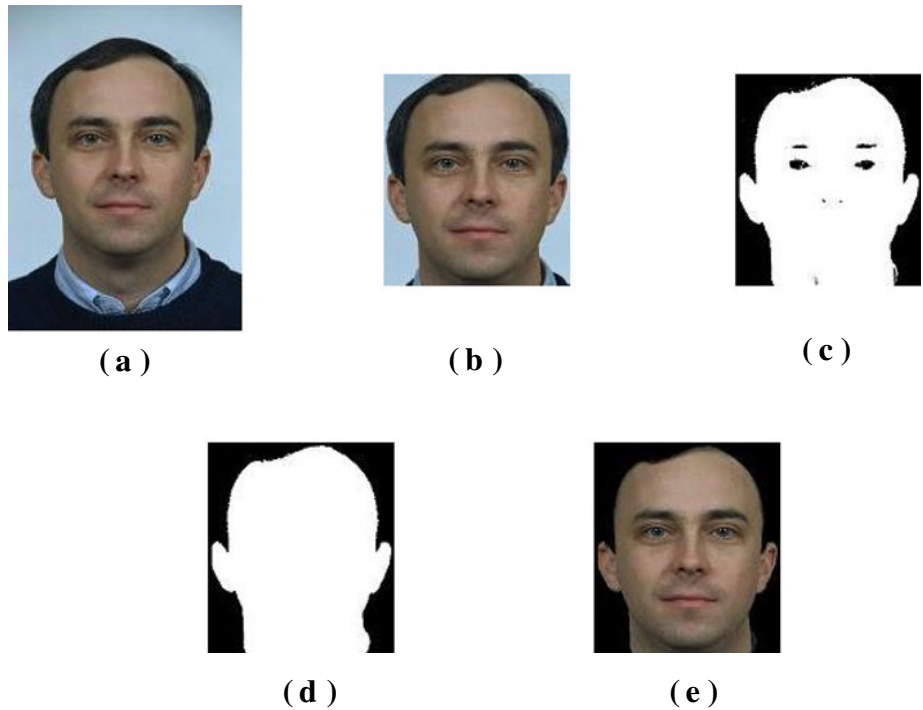

**Figure 5 face localization and segmentation (a)** A face image from FERET face database, **(b)** the segmented image by using local SMQT, **(c)** the mask generated by applying the skin color-based face detection [14], **(d)** the mask obtained by hole filling of the previous mask, and **(e)** the final segmented face image.

system [14]. The proposed combination of the aforementioned face segmentation techniques gives an outline of Figure 5, in which a face image from the FERET face database is entered into the system.

## PDF based face recognition by using LBP
### Analytical point of view
In a general mathematical sense, an image PDF is simply a mapping $\eta_i$ representing the probability of the pixel intensity levels that fall into various disjoint intervals, known as bins. The bin size determines the size of the PDF vector. Given a monochrome image, PDF $\eta_j$ meets the following conditions, where $N$ is the total number of pixels in an image and the bin size is 256:

$$N = \sum_{j=0}^{255} \eta_j \tag{8}$$

Then, PDF feature vector, $H$, is defined by

$$H = [p_0, p_1, ..., p_{255}]$$
$$p_\iota = \frac{\eta_\iota}{N} \quad , \quad \iota = 0, ..., 255 \tag{9}$$

There are two issues in this point. First one is which metric is more suitable for classification of the PDFs of the faces. For this purpose, L1, L2, cross correlation, and Kullback–Leibler distance (KLD) have been used for the classification of gray-scale faces in the HP face database. The recognition rate is reported in Table 1. It shows that KLD performs better than the other metrics. Hence in this study, similar to Demirel and Anbarjafari's study [6,8], we use KLD as a metric for finding the divergence between the PDFs.

The second issue is the number of bin size. What will happen if the bin size drops from 256 into the smaller value? In line of this issue, the bin sizes have been changed and the recognition rate for the gray-scale face images in the HP face database has been obtained. Table 2 reports this result which shows the performance drops as the bin size decreases, due to loss of information laid on the pattern of the PDF.

**Table 1 Recognition rate performance (%) of the PDF-based face recognition system for gray-scale PDFs of HP face database with 15 subjects and 10 samples per each subject obtained by using four different metrics**

| # of Training Images | L1 | L2 | Cross Correlation | KLD |
|---|---|---|---|---|
| 1 | 48.63 | 36.48 | 50.81 | 54.81 |
| 2 | 57.92 | 48.5 | 69.54 | 78.00 |
| 3 | 70.71 | 61.76 | 77.6 | 84.10 |
| 4 | 77.78 | 71.78 | 82.26 | 88.00 |
| 5 | 76.73 | 67.8 | 87.78 | 93.87 |

**Table 2 Recognition of face images of the HP face database by using different bin numbers**

| Number of bins | Recognition rates (%) |
|---|---|
| 256 | 96.00 |
| 128 | 93.33 |
| 64 | 90.67 |
| 32 | 82.67 |

One of the missing studies in Demirel and Anbarjafari's work was analysis of the discrimination power of the PDF. In order to show the discrimination power of the PDFs by using KLD, the average KLD value within and between-class distances of various databases in different color spaces are given in Table 3.

In Table 3 class discrimination, $\varnothing_c$, is defined to be the ratio of the average between-class distance and the average within class distance which is indicating the discrimination power of different color channels. Class discrimination values show that KLD provides enough separation between classes in different color channels in PDF-based face recognition.

As reported in [6,8], the PDF-based face recognition system can be implemented in various color channels such as *HSI* and *YCbCr* color spaces in which the luminance and chrominance are separated from each other. These multi decisions can be combined later by using various fusion techniques. But as some of these color channels are highly correlated with each other (they have high mutual entropy), there is no need to include the repeated information in decision making. Table 4 shows the mutual entropy between the channels of *HSI* color space and the *YCbCr* color space.

**Table 3 The discrimination of the PDF by using KLD in different color channels and different databases**

| DATA BASE | | AVERAGE WITHIN CLASS DISTANCE | AVERAGE BETWEEN CLASS DISTANCE | CLASS DISCRIMINATION $\varnothing_c$ |
|---|---|---|---|---|
| FERET | H | 0.0542 | 0.3316 | 6.12 |
| | S | 0.0909 | 0.4008 | 4.41 |
| | I | 0.0716 | 0.2396 | 3.35 |
| | Y | 0.0260 | 0.1300 | 5.00 |
| | Cb | 0.0435 | 0.1689 | 3.88 |
| | Cr | 0.0364 | 0.1335 | 3.67 |
| HP | H | 0.0159 | 0.0818 | 5.14 |
| | S | 0.0097 | 0.0634 | 6.54 |
| | I | 0.0141 | 0.0612 | 4.34 |
| | Y | 0.0141 | 0.0611 | 4.33 |
| | Cb | 0.0132 | 0.0582 | 4.41 |
| | Cr | 0.0126 | 0.0573 | 4.55 |
| BOS | H | 0.0083 | 0.2166 | 26.11 |
| | S | 0.0093 | 0.9850 | 106.38 |
| | I | 0.0334 | 0.2793 | 8.36 |
| | Y | 0.0336 | 0.3385 | 10.07 |
| | Cb | 0.0209 | 0.6007 | 28.74 |
| | Cr | 0.0243 | 0.5027 | 20.70 |

Table 4 shows the average mutual information in percentage between the various color channels in *HSI* and *YCbCr* color spaces for Bosphorus face databases where there exist over 4,500 face images. The high correlation between *I-Y*, *I-Cb*, and *I-Cr* color channels shows that instead of using both color spaces, using only *HSI* will have enough information in order to get conclusive recognition rate after the fusion. Also Table 4 indicates that the color channels in *YCbCr* are highly correlated with each other.

## Local binary pattern

The local binary pattern (LBP) is a non-parametric operator which describes the local spatial structure of an image [12,41]. Ojala et al. [41] introduced this operator and showed its high discriminative power for texture classification. At a given pixel position $(x,y)$, LBP is defined as an ordered set of binary comparisons of pixel intensities between the center pixel and its eight neighbor pixels, as shown in Figure 6.

The decimal form of the resulting 8-bit word of LBP code can be expressed as follows.

$$\text{LBP}(x,y) = \sum_{n=0}^{7} 2^n s\left(i_n - i_{(x,y)}\right) \tag{10}$$

where $i_{(x,y)}$ corresponds to the gray value of the center pixel $(x,y)$, $i_n$ to the gray values of the eight neighbor pixels, and function $s(x)$ is defined as

$$s(x) = \begin{cases} 1 \ if \ x \geq 0 \\ 0 \ if \ x < 0 \end{cases} \tag{11}$$

By definition, the LBP operator is unaffected by any monotonic gray-scale transformation which preserves the pixel intensity order in a local neighborhood. Note that each bit of the LBP code has the same significance level and that two successive bit values may have a totally

**Table 4 The correlation between *HSI* and *YCbCr* color channels in percentage**

| | H | S | I | Y | Cb | Cr |
|---|---|---|---|---|---|---|
| **H** | 100 | 26.49 | 9.36 | 5.28 | 9.29 | 16.13 |
| **S** | 26.49 | 100 | 25.97 | 32.91 | 52.51 | 65.58 |
| **I** | 9.36 | 25.97 | 100 | 99.05 | 98.32 | 93.39 |
| **Y** | 5.28 | 32.91 | 99.05 | 100 | 96.02 | 84.63 |
| **Cb** | 9.29 | 52.51 | 98.32 | 96.02 | 100 | 92.50 |
| **Cr** | 16.13 | 65.58 | 93.39 | 84.63 | 92.50 | 100 |

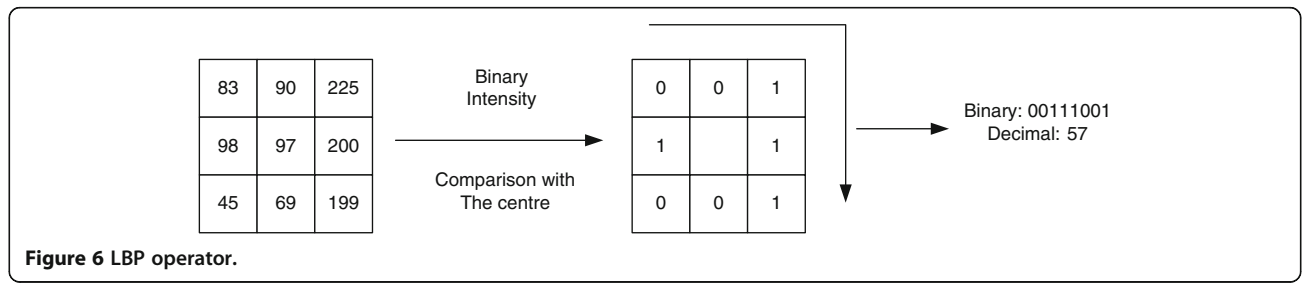

**Figure 6 LBP operator.**

different meaning. Sometimes, the LBP code is referred as a kernel structure index.

Ojala et al. [42] extended their previous study to a circular neighborhood of different radius size. They used $LBP_{P,R}$ notation which refers to $P$ equally spaced pixels on a circle of radius $R$. Two of the main motivations of using LBP are its low computational complexity and its texture discriminative property. LBP has been used in many image processing applications such as motion detection [43], visual inspection [44], image retrieval [45], face detection [46], and face recognition [47,48].

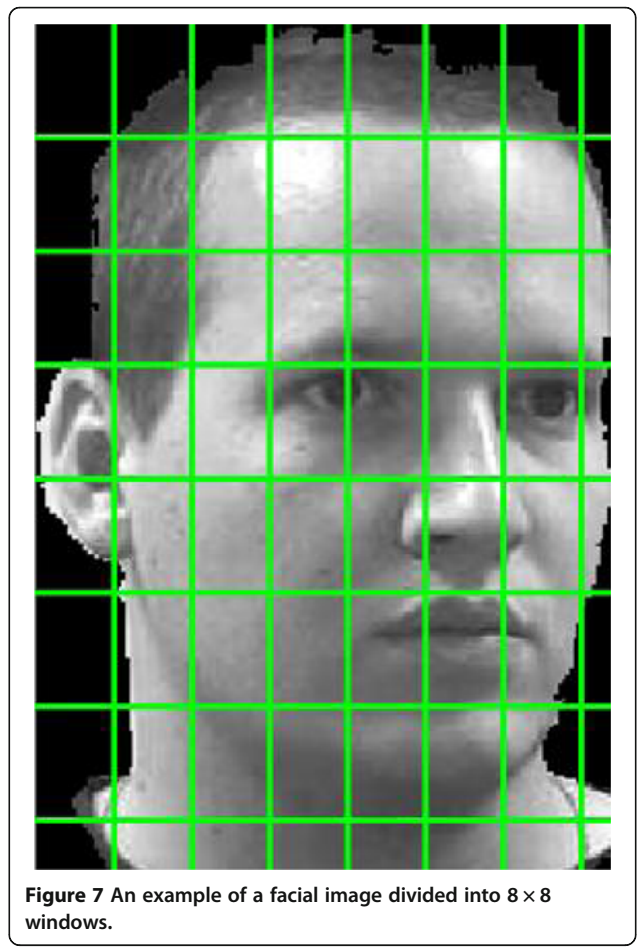

**Figure 7 An example of a facial image divided into 8 × 8 windows.**

In most aforementioned applications, a face image was usually divided into small regions. For each region, a cumulative histogram of LBP code computed at each pixel location within the region was used as a feature vector.

Ahnon et al. [48] used LBP operator for face recognition. Their face recognition system can be explained as follows: a histogram of the labeled image $f_1(x,y)$ can be defined as

$$Hi = \sum_{x,y} I\{f1(x,y) = i\} \quad i = 0, \cdots, n-1 \quad (12)$$

where $n$ is the number of different labels produced by the LBP operator and

$$I\{A\} = \begin{cases} 1 & A \text{ is true} \\ 0 & A \text{ is false} \end{cases} \quad (13)$$

This histogram contains information about the distribution of the local micropatterns, such as edges, spots, and flat areas, over the whole image. For efficient face representation, retaining the spatial information is required; hence, the image is divided into regions $R_0$, $R_1$, ..., $R_{m-1}$, as shown in Figure 7.

The spatially enhanced histogram is defined as

$$H_{i,j} = \sum_{x,y} I\{f_1(x,y) = i\} I\{(x,y) \in R_j\} \quad , \quad \begin{matrix} i = 0, \cdots, n-1 \\ j = 0, \cdots, m-1 \end{matrix}$$
$$(14)$$

In this histogram, a description of the face on three different levels of locality exists: the labels for the histogram contain information about the patterns on a pixel level, the labels are summed over a small region to produce information on a regional level, and the regional histograms are concatenated to build a global description of the face.

Although Ahnon et al. [48] have mentioned several dissimilarity measures such as histogram intersections, log-likelihood statistics, and Chi square statistics, they used nearest neighbor classifier in their study.

When the image has been divided into several regions, it can be expected that some of the regions contain more useful information than others in terms of distinguishing between people, such as eyes [49,50]. In order to contribute such information, a weight can be set for each region based on the level of information it contains.

**Table 5 Performance of different decision-making techniques for the proposed face recognition system**

| | # of training image | SUM RULE | MEDIAN RULE | MV | FVF |
|---|---|---|---|---|---|
| **H** | 1 | 82.82 | 83.37 | 80.36 | 79.94 |
| **P** | 2 | 95.68 | 95.55 | 91.75 | 86.09 |
| | 3 | 96.45 | 96.31 | 92.93 | 95.25 |
| | 4 | 96.84 | 97.14 | 95.02 | 96.85 |
| | 5 | 98.08 | 98.38 | 98.70 | 97.32 |
| **F** | 1 | 76.63 | 75.90 | 75.09 | 83.32 |
| **E** | 2 | 86.49 | 87.74 | 87.79 | 89.43 |
| **R** | 3 | 90.76 | 89.41 | 87.72 | 97.13 |
| **E** | 4 | 95.81 | 93.56 | 92.00 | 99.01 |
| **T** | 5 | 96.36 | 97.17 | 94.34 | 99.78 |
| **B** | 1 | 74.85 | 74.94 | 77.54 | 79.04 |
| **O** | 7 | 88.60 | 88.44 | 88.63 | 89.81 |
| **S** | 13 | 89.97 | 89.32 | 89.77 | 90.10 |
| | 19 | 90.07 | 92.04 | 93.01 | 93.12 |
| | 25 | 91.91 | 94.07 | 94.31 | 95.67 |

**The proposed LBP-based face recognition**

In this article, the proposed face recognition system uses LBP in order to obtain different PDF of each face in different color channels. Each face images after being equalized and segmented will be divided into sub-images in different color channels. For each sub-image in a specific color channel, the PDF will be calculated. The concatenation of these PDFs will result into a single PDF. Due to high correlation of information of $I$ color channel with $YCbCr$ color channels, only PDFs of $HSI$ color space is used for recognition process by using KLD as it was mentioned earlier. Because several decisions have been obtained from different color channels, the combination of these decisions will boost the final decision.

**Table 6 Performance of the proposed LBP based face recognition system using FVF, PCA, LDA, conventional gray scale LBP, PDF based face recognition, NMF, and INMF based face recognition system for the FERET face databases with 50 subjects and 10 samples per each subject**

| # of training images | 1 | 2 | 3 | 4 | 5 |
|---|---|---|---|---|---|
| **FVF** | 83.32 | 89.43 | 97.13 | 99.01 | 99.78 |
| **FVF** [8] | 82.89 | 87.00 | 96.57 | 98.80 | 99.33 |
| **PCA** | 44.00 | 52.00 | 58.29 | 66.17 | 68.80 |
| **LDA** | 61.98 | 70.33 | 77.78 | 81.43 | 85.00 |
| **LBP** | 50.89 | 56.25 | 74.57 | 77.67 | 79.60 |
| **NMF** | 61.33 | 64.67 | 69.89 | 77.35 | 80.37 |
| **INMF** | 63.65 | 67.87 | 75.83 | 80.07 | 83.20 |

In [51], various decision fusion, data fusion, and source fusion techniques have been studied. In this study, sum rule, median rule, majority voting, and feature vector fusion (obtained by concatenating the PDFs before starting the recognition process) have been used in order to combine the multi decisions obtained from LBP–PDF-based face recognition in $HSI$ color channels. Table 5 shows the correct recognition of the aforementioned fusion techniques for the HP face database with 15 subjects and 10 samples per each subject, FERET face database with 50 subjects and 10 samples per each subject, and Bosphorus face database with 105 subjects and 32 samples per each subject.

In order to show the superiority of the proposed method on available state-of-the-art and conventional face recognition systems, we have compared the recognition performance with conventional PCA-based face recognition system and the state-of-the-art techniques such as, NMF [52,53], supervised INMF [54], conventional gray-scale LBP [45], and LDA-based face recognition systems [3] for the FERET face database. The experimental results are shown in Table 6. The results clearly indicate that this superiority is achieved by using PDF-based face recognition in different color channels backed by the data fusion techniques. Compared to the state-of-the-art technique proposed by Demirel and Anbarjafari [8], the recognition rate obtained by using LBP has slightly better performance.

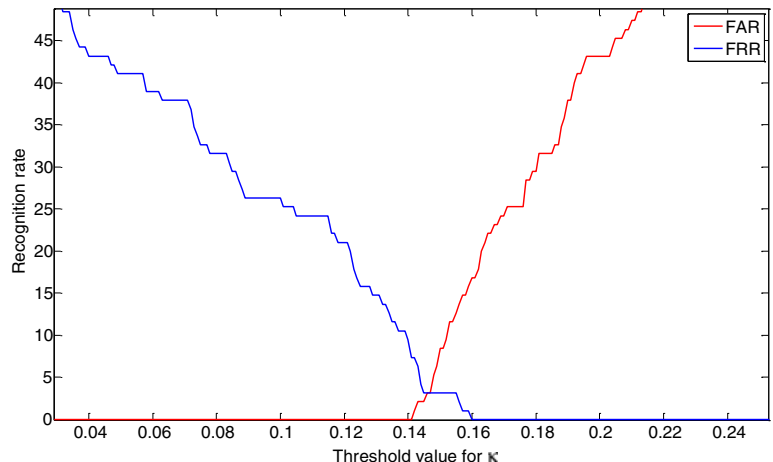

**Figure 8 FAR and FRR curves.**

Figure 8 shows the FAR and FRR analysis of the proposed face recognition system for the FERET face database. The equal error rate occurs when the recognition rate is about 3.2% which shows that the system can be used in a practical scenario.

The median rule and FVF-based results are 97.17 and 99.78% for the FERET face database, when five samples per subject are available in the training set, respectively. These results are significant, when compared with the recognition rates achieved by conventional PCA and LDA and the state-of-the-art techniques such as LBP, NMF, and INMF-based face recognition system.

## Conclusion
In this article, we have studied a high performance frontal homogenous illumination robust face recognition system using LBP and PDFs in different mutually independent color channels. A face localization which is a combination of local SMQT technique followed by skin tone-based face detection method was employed in this study. DWT + SVD-based image illumination enhancement technique was also applied in order to reduce the effect of illumination. The article analytically analyzed and justified the use of KLD and PDF with a bin size of 256 which was introduced and used in [6,8]. Several well-known fusion techniques have been used in order to combine the decisions obtained from mutually independent color channels. Also an FAR and FRR analysis has been done in this study. Finally, comparison between the proposed method and the conventional and the state-of-the-art techniques has been done which showed the superiority of the proposed method.

### Competing interests
The authors declare that they have no competing interests.

### Acknowledgments
The author would like to thank Asst. Prof. Dr. Mikael Nilsson from Blekinge Institute of Technology, for providing the algorithm for local SMQT-based face recognition. Also the author would like to thank Prof. Dr. Ivan Selesnick from Polytechnic University for providing the DWT codes in MATLAB.

Received: 19 September 2012 Accepted: 1 November 2012
Published: 22 January 2013